\documentclass[10pt,twocolumn,letterpaper]{article}

\usepackage{iccv}
\usepackage{times}
\usepackage{epsfig}
\usepackage{graphicx}
\usepackage{amsmath}
\usepackage{amssymb}

\usepackage{graphics} 
\usepackage[ruled,vlined]{algorithm2e}
\usepackage{caption}
\usepackage{subcaption}
\SetKw{kwinput}{Input}
\SetKw{kwoutput}{Output}

\usepackage[pagebackref=true,breaklinks=true,letterpaper=true,colorlinks,bookmarks=false]{hyperref}

\iccvfinalcopy 

\def\iccvPaperID{10806} 

\ificcvfinal\fi

\begin{document}
\newcommand{\xiaowei}[1]{{\color{blue}#1}}
\newcommand{\dengyu}[1]{{\color{red}#1}}
\newcommand{\xy}[1]{{\color{cyan}#1}}

\newcommand{\commentout}[1]{}
\newcommand{\CAF}{\kappa}
\newcommand{\TSF}{\alpha}

\newtheorem{proposition}{Proposition}
\newtheorem{remark}{Remark}
\title{A Little Energy Goes a Long Way: Build an Energy-Efficient, Accurate Spiking Neural Network from Convolutional Neural Network}

\author{Dengyu Wu, Xinping Yi, and Xiaowei Huang\\
University of Liverpool, UK\\
{\tt\small \{dengyu.wu,xinping.yi,xiaowei.huang\}@liverpool.ac.uk } 
}

\maketitle
\ificcvfinal\thispagestyle{empty}\fi

\begin{abstract}
This paper conforms to a recent trend of developing an energy-efficient Spiking Neural Network (SNN), which takes advantage of the sophisticated training regime of Convolutional Neural Network (CNN) and converts a well-trained CNN to an SNN. We observe that the existing CNN-to-SNN conversion algorithms may keep a certain amount of residual current in the spiking neurons in SNN, and the residual current may cause significant accuracy loss when inference time is short. To deal with this, we propose a unified framework to equalise the output of convolutional or dense layer in CNN and the accumulated current in SNN, and maximally align spiking rate of a neuron with its corresponding charge. This framework enables us to design a novel explicit current control (ECC) method for the CNN-to-SNN conversion which considers multiple objectives at the same time during the conversion, including accuracy, latency and energy efficiency. We conduct an extensive set of experiments on different neural network architectures, e.g. VGG, ResNet and DenseNet, to evaluate the resulting SNNs. The benchmark datasets include not only the image datasets such as CIFAR-10/100 and ImageNet but also the Dynamic Vision Sensor (DVS) image datasets such as DVS-CIFAR-10. The experimental results show the superior performance of our ECC method over the state-of-the-art.

\end{abstract}

\section{Introduction}


Spiking neural networks (SNNs) 
are more energy efficient than convolutional neural networks (CNNs) in inference time thanks to its utilisation of matrix addition instead of multiplication. SNNs are supported by new computing paradigms and hardware. For example,
SpiNNaker \cite{Spinnaker:2012}, a neuromorphic computing platform based on SNNs, can run real-time billions of neurons to simulate human brain. The neuromorphic chips, such as TrueNorth \cite{TrueNorth:2015}, Loihi \cite{Loihi:2018}, and Tianji \cite{Pei:2019}, 
can directly implement SNNs with 
10,000 neurons being integrated onto a single chip. Moreover, through the combination with sensors,  SNNs can be applied to edge computing, robotics, and other fields, to build low-power intelligent systems \cite{challenges:2018}.

However, 
the discrete  nature of spikes makes the training of SNNs hard, due to the absence of gradients. 
%
%
\begin{figure*}[h!]
    \centering
    \vspace{-5pt}
    \resizebox{\textwidth}{!}{
    \includegraphics{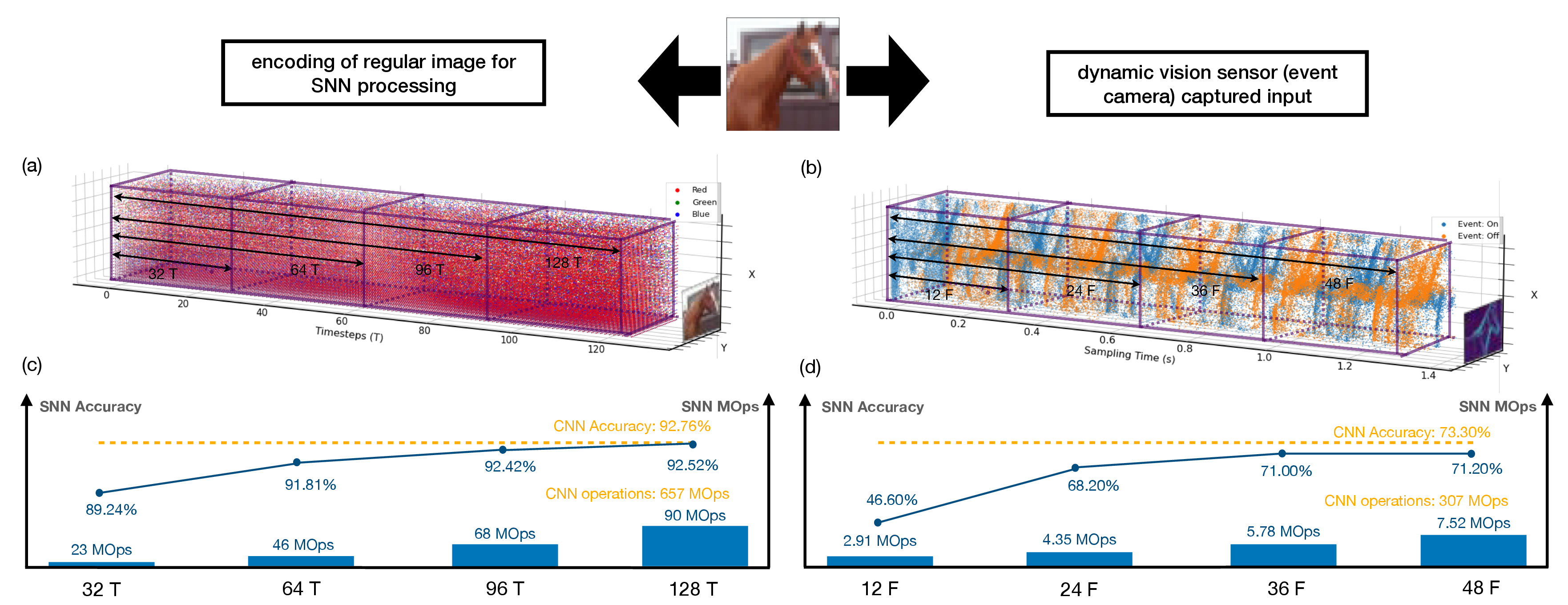}
    }
    \vspace{-5pt}
    \caption{An illustrative diagram showing how SNNs process two different types of inputs and their performance comparison with CNNs. A regular image (left column) -- taken from camera -- is preprocessed into a spike train (A), which  then runs through the SNN in a number of timesteps (e.g., 128 timesteps as in the figure). A DVS input -- taken from event camera -- can be represented directly as a spike train (B), and processed naturally by the SNN in a number of frames (e.g., 48 frames as in the figure). 
    (C) and (D) show the SNN's performance with respect to the three objectives (accuracy, energy efficiency, and latency), comparing to CNNs. }
        \vspace{-5pt}
    \label{fig:introductionDescription}
\end{figure*}
%
This paper follows a cutting-edge approach of obtaining a well-performed SNN by 
converting from a trained CNN 
of the same structure. This approach has an obvious benefit from the sophisticated training regime of CNNs, i.e., it is able to take advantage of the successful -- and still fast improving -- training methods on CNNs without extra efforts to adapt them to SNNs.  Unfortunately,  
existing CNN-to-SNN conversion methods either cannot achieve a sufficiently small accuracy loss upon conversion \cite{Rueckauer:2017,Sengupta:2019}, or need a high latency
\cite{Sengupta:2019},  or require a significant increase on the energy consumption of the resulting SNNs \cite{Han:2020}. Moreover, recent methods such as \cite{Han:2020} do not work with batch-normalisation layer -- a functional layer that plays a key role in the training of CNNs \cite{DBLP:conf/nips/SanturkarTIM18}.

This paper levels up the CNN-to-SNN conversion with the following contributions. First of all, methodologically, we argue that, the conversion needs to be multi-objective --  in addition to accuracy loss, energy efficiency and latency should be considered altogether. 
%
Figure~\ref{fig:introductionDescription} provides an illustration showing 
how SNNs process images and DVS inputs, exhibiting how well our methods enable the achievement of the three objectives and its comparison with CNNs. Actually, from (C) and (D) of  Figure~\ref{fig:introductionDescription}, 
SNNs can have 
competitive accuracy
upon conversion (92.52\% vs 92.76\%, and 71.20\% vs 73.30\%, respectively) and be significantly more energy efficient than CNNs (90MOps vs 657MOps, and 7.52MOps vs 307MOps, respectively). 
While it is hard to compare the latency as SNNs and CNNs work on different settings, our method implements the 
high energy efficiency with low latency (128 timesteps for images and 48 frames for DVS inputs). As shown in our experiments, ours are superior to the state-of-the-art conversions \cite{Rueckauer:2017,Sengupta:2019,Han:2020}.

Second, we follow an intuitive view  aiming to  
establish an equivalence between the activations in an original CNN and the current in the resulting SNN. This view 
inspires us to consider  
an explicit, and detailed, control on the current flowing through the SNN.
Technically, we 
develop a unifying theoretical framework, which treats both weight normalisation \cite{Rueckauer:2017} and threshold balancing \cite{Sengupta:2019} as special cases. Based on the framework, we develop a novel conversion method called explicit current control (ECC), which includes two techniques:   current normalisation (CN),  to control the maximum number of spikes fed into the SNN, and thresholding for residual elimination (TRE), to reduce the residual membranes potential in the neurons. 

Third, 
we include in ECC a dedicated technique called consistency maintenance for batch-normalisation (CMB) to deal with the conversion of batch-normalisation layer.




Finally, we implement ECC into a tool SpKeras\footnote{https://github.com/Dengyu-Wu/spkeras} and conduct an extensive set of experiments on not only the regular image datasets, such as CIFAR-10/100 and ImageNet, but also the Dynamic Vision Sensor (DVS) datasets such as DVS-CIFAR-10. Note that, DVS datasets are dedicated for SNN processing. 
The experimental results show that, comparing with state-of-the-art methods \cite{Rueckauer:2017,Sengupta:2019,Han:2020}, ECC can optimise over three objectives at the same time, and have superior performance. Moreover, we notice that (1) ECC can  utilise the conversion of batch-normalisation to reduce the latency, and (2) ECC is robust to the hardware deployment because the quantisation -- by using 7-10 bits to represent the originally 32-bit weights -- does not lead to significant accuracy loss.   

We remark that, this paper is not to argue for the replacement of CNNs with SNNs in general. Instead, we suggest a plausible deployment workflow, i.e., train a CNN $\rightarrow$ convert into an SNN $\rightarrow$ deploy on edge devices with e.g., event camera. 
The workflow will not be a good option if any of the three objectives is not optimised.  



\section{Related Work}

\subsection{Current ``energy for accuracy'' trend in CNN-to-SNN Conversion}

A few different conversion methods, such as \cite{Diehl:2015,Rueckauer:2017,Sengupta:2019,Han:2020}, have been proposed in the past few years. It is not surprising 
that there is an accuracy loss between SNNs and CNNs. For example, in \cite{Rueckauer:2017,Sengupta:2019}, the gap is between 0.15\% - 2\% for CIFAR10 networks. A recent work \cite{Sengupta:2019} shows that this gap can be reduced if we use a sufficiently long (e.g., 1024 timesteps) spike train to encode an 
input. However, a longer spike train will inevitably lead to higher latency.
This situation was believed to be eased in \cite{Han:2020}, which claims that the length of spike train can be drastically shortened in order to achieve near-zero accuracy loss. However, as shown in Section~\ref{sec:TMTScomparison} (Figure~\ref{fig:TMTScomparison}A), their threshold scaling method can easily lead to a significant increase on the spike-caused synaptic operations \cite{Rueckauer:2017}, or spike operations for short, which also lead to significant increase on the energy consumption. 

Other related works include \cite{Rathi2020Enabling,rathi2021diet,pmlr-v139-li21d}, which calibrate SNN to a specific timestep by gradient-based optimisation. The calibration requires extra training time to find the optimal weights or hyper-parameters, such as some thresholds. In contrast, \cite{deng2021optimal} trained a dedicated CNN for an SNN with fixed timesteps by shifting and 
clipping ReLU activations, although the accuracy loss of these SNNs cannot converge to zero when increasing the timesteps, as shown in Figure \ref{fig:clippedmethods}B. Besides, instead of reducing spikes, \cite{lu2020exploring} explored SNN with binary weights to further improve the energy efficiency by consuming less memory.
\subsection{Technical Ingredients in CNN-to-SNN Conversion}
\begin{table*}[]
    \centering
    \begin{tabular}{|l||c|c|c|c|c|c||c|c|c|}
    \hline
         & HR & SR & WN$^{*}$& TB$^{*}$ & TS  & ECC & BN$^{**}$ & MP & AP\\
         \hline

        \cite{Cao:2015} &$\surd$ &  &   & &  &  &  &  & $\surd$\\

        \cite{Diehl:2015} &$\surd$ &  &$\surd$   & &  &  &  &  &$\surd$ \\
         
        \cite{Rueckauer:2017} & & $\surd$ & $\surd$  & &  &  & $\surd$ & $\surd$ & \\
        
        \cite{Sengupta:2019} & $\surd$ &  & & $\surd$ & & & & & $\surd$ \\
        
        \cite{Han:2020} &  & $\surd$ & &  $\surd$ & $\surd$ &   & & & $\surd$ \\
        
         [this paper] &  & $\surd$ &  $\surd$ &  $\surd$ & & $\surd$ & $\surd$ & & $\surd$\\
        \hline
     \end{tabular}
    \caption{Comparison of Key Technical Ingredients (HR, SR, WN, TB, TS, ECC) and Workable Layers (BN, MP, AP) with the State-of-the-Art Methods. HR: hard reset; SR: reset by subtraction, or soft reset; WN: weight normalisation; TB: threshold balancing; TS: threshold scaling;  ECC: explicit current control; BN: batch normalisation; MP: max pooling; AP: average pooling. * As a contribution of this paper, in Section~\ref{sec:TBWN}, we show that both WN and TB are special cases of our ECC framework. ** Among all methods, only those that can handle BN have bias terms in their pre-trained CNNs. }
    \label{tab:theoreticalcomparison}
\end{table*}
Table~\ref{tab:theoreticalcomparison} provides an overview of the existing conversion methods \cite{Cao:2015,Diehl:2015,Rueckauer:2017,Sengupta:2019,Han:2020} and ours, from the aspects of technical ingredients and workable layers. At the beginning, most of the techniques, such as \cite{Cao:2015,Diehl:2015}, are based on hard reset (HR) spiking neurons, which are reset to fixed reset potential once its membrane potential exceeds the firing threshold. HR is still used in some recent methods such as \cite{Sengupta:2019}. The main criticism on HR is its significant information loss during the SNN inference. Soft reset (SR) neurons 
are shown better in other works such as \cite{Rueckauer:2017,Han:2020}.

Weight normalisation (WN) is proposed in \cite{Diehl:2015} and extended in \cite{Rueckauer:2017} to regulate the spiking rate in order to reduce accuracy loss. The other technique, threshold balancing (TB), is proposed in \cite{Sengupta:2019} and extended in \cite{Han:2020}, to assign appropriate threshold to the spiking neurons to ensure that they operate in the linear (or almost linear) regime. 
We show in Section~\ref{sec:TBWN} 
that both WN and TB are special cases of our theoretical framework. 

Another technique, called threshold scaling (TS), is suggested in \cite{Han:2020}. However, as our experimental result shown in Figure~\ref{fig:TMTScomparison}A, 
TS leads to a significantly greater energy consumption (measured as MOps). On the other hand, 
our ECC method can achieve smaller accuracy loss and significantly less energy consumption.

We also note in Table~\ref{tab:theoreticalcomparison} the differences 
in terms of workable layers 
in CNNs/SNNs 
for different methods.
For example, batch-normalisation (BN) layer \cite{pmlr-v37-ioffe15} is known important for the optimisation of CNNs \cite{DBLP:conf/nips/SanturkarTIM18}, but only one existing method, i.e., \cite{Rueckauer:2017}, can work with it. Similarly for the bias values of neurons which are pervasive for CNNs. Actually, the consideration of BN is  argued in \cite{Sengupta:2019} as the key reason for the higher accuracy loss in \cite{Rueckauer:2017}.  The results of this paper show that, we can keep both 
 BN and bias 
 without significant increased energy consumption, by maintaining the consistency between the behaviour of SNN and CNN. BN can help with the reduction of latency. As we discussed earlier and in Section 5, our ECC method may be applicable to B-SNN and further improve its performance. 
%
Moreover, we follow most  SNN research to consider 
average pooling (AP) layer instead of max pooing (MP) layer. 

\subsection{Direct Training} 

SNNs process information through non-differentiable spikes, and thus the 
backpropagation (BP) \cite{BP:1989} training algorithm cannot be directly applied.
A few attempts \cite{Lee:2016,dtraining:2020} have been made to adapt the BP algorithm by approximating its forward propagation phase.
Such direct training  requires high computational complexity to achieve an accuracy that is close to CNNs \cite{wu2021training}. 
Unlike these methods which approximate the BP algorithm \cite{Lee:2016,dtraining:2020}, both of which may lead to performance degradation, we choose CNN-to-SNN conversion which can take full advantage of the continuously improving CNN training methods. Other than these methods which try to re-produce the success of CNN training, there are other direct training methods, such as approaches based on reservoir computing (\cite{soures2019spiking}) and evolutionary algorithms   (\cite{schuman2020evolutionary}).

\section{Explicit Current Control (ECC)}\label{sec:methodology}

By leveraging the correspondence between activation in CNNs and current in SNNs,\footnote{The activation values in the original CNNs can be represented by the current through the analog/digital circuits in the resulting SNNs, so that controlling current through spike train in SNNs corresponds to data flow operations in CNNs.}
we propose a unifying theoretical framework targeting multiple objectives, including accuracy, latency and energy efficiency. Going beyond the existing conversion techniques (see Table \ref{tab:theoreticalcomparison}) that consider some of the objectives individually, we view these multi-objective holistically through the lens of the unifying theoretical framework. 
Inspired by such new viewpoint, we develop explicit current control (ECC) techniques to normalise, clip, and maintain the current through the SNNs for the purposes of reducing accuracy loss, latency, and energy consumption. 
\subsection{Existing CNN-to-SNN Conversion}

%

Without loss of generality, we consider 
a CNN model of $N$ layers such that layer $n$ has $M^n$ neurons, for $n \in \{1,2,\dots,N\}$.
The output of the neuron $i\in\{1,\dots,M^n\}$ at layer $n$ with ReLU activation function is given by
\begin{align}
a_i^n = \max \left\{0, \sum_{j=1}^{M^{n-1}} W_{ij}^n a_{j}^{n-1}  + b^n_i \right\}
\end{align}
where $W_{ij}^n$ is the weight between the neuron $j$ at layer $n-1$ and the neuron $i$ at layer $n$, $b_i^n$ is the bias of the neuron $i$ at layer $n$, and $a_i^0$ is initialised as the input $x_i$. 

The activation $a_i^n$ indicates the contribution of the neuron to the CNN inference. 
For CNN-to-SNN conversion, the greater $a_i^n$ is, the higher spiking rate will be, for the corresponding neuron
on SNN. 
%
An 
explanation of a conversion method from CNNs to SNNs was first introduced in \cite{Rueckauer:2017} by using data-based weight normalisation. 

The conversion method uses integrated-and-fire (IF) neuron to construct a rate-based SNN without leak and refractory time. If considering practical implementations, the rate-based SNN expects a relatively large interval between input spikes to minimise the effect of refractory time. To convert from a CNN, the spiking rate of each neuron in SNN is related to the activation of its corresponding neuron in the CNN. An iterative algorithm based on the \emph{reset by subtraction} mechanism is described below. The membrane potential $V_{i}^n(t)$ 
of the neuron $i$ at the layer $n$ can be described as

\begin{align} \label{eq:em_potential}
    V_{i}^n(t) = V_{i}^n(t-1) + Z_i^n(t) - \Theta_{i}^n (t)V^n_{thr}
\end{align}
where $V^n_{thr}$ represents the threshold value of  layer $n$ and  $Z^n_i(t)$ is the input current to neuron $i$ at layer $n$ such that 
\begin{align}
    Z^n_i(t) = \sum_{j=1}^{M^{n-1}} W_{ij}^n \Theta_{j}^{n-1} (t)  + b^n_i
\end{align}
with $\Theta_{i}^n (t)$ being a step function defined as
\begin{align}\label{equ:originaltheta}
    \Theta_{i}^n (t) = \left\{ \begin{matrix} 1,  &\text{ if  } V_{i}^n(t)  \geq  V_{thr}^n \\ 0, & \text{otherwise.     } \end{matrix} \right. 
\end{align}
In particular, when the current $V^n_i(t)$ reaches the threshold $V_{thr}^n$, the neuron $i$ at layer $n$ will generate a spike, indicated by the step function $\Theta_{i}^n (t)$, and the membrane potential $V_{i}^n(t)$ will be reset immediately for the next timestep by subtracting the threshold. 

\subsection{A Unifying Theoretical Framework}
\label{sec:TBWN}

\begin{figure*}[t]
    \centering
    \resizebox{\textwidth}{!}{
    \includegraphics{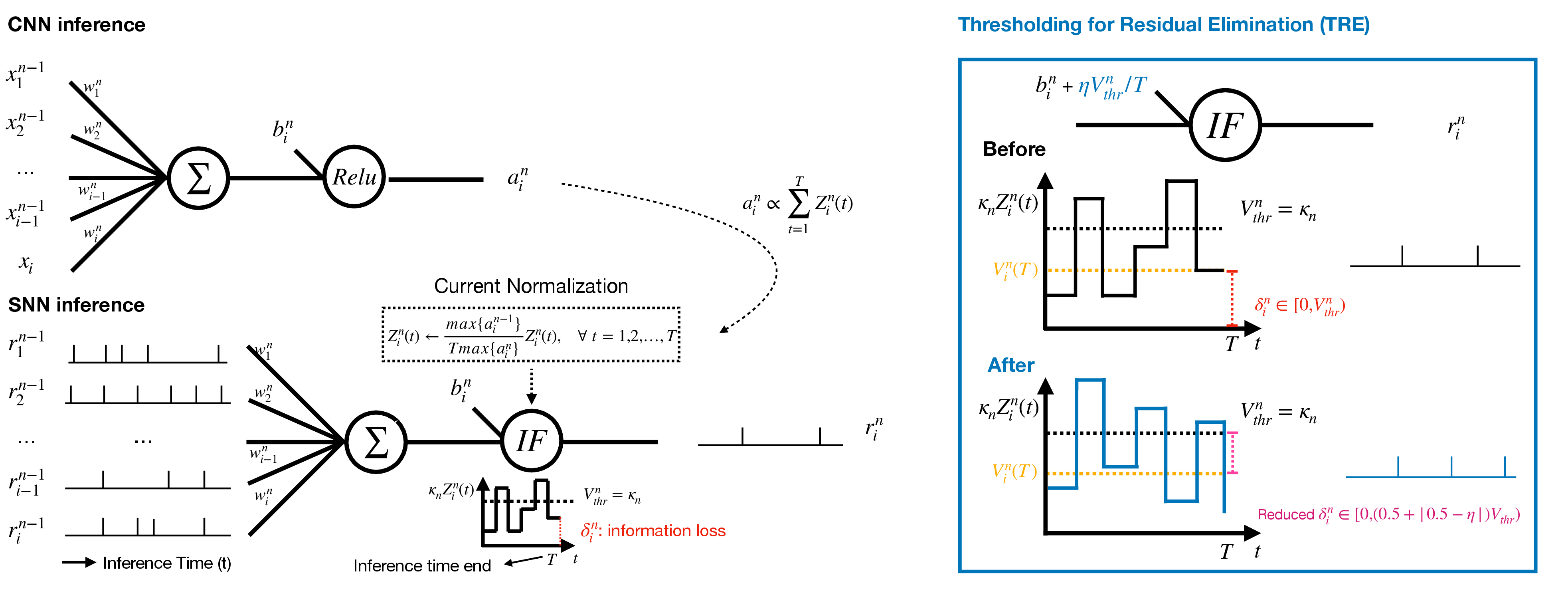}
    }
    \vspace{-10pt}
    \caption{{\textbf{Left}: Our proposed CNN-to-SNN conversion for the $n$-th layer with a current normalisation component and a thresholding mechanism. The activation $a_i^n$ in the CNN (Top) is used for current normalisation in the SNN (Bottom). \textbf{Right}: The proposed Thresholding for Residual Elimination (TRE) and the illustration of error reduction by TRE.
    }}
    \label{fig:CNN-SNN-conv}
\end{figure*}

The above 
CNN-to-SNN conversion method is designed specifically for 
weight normalisation \cite{Rueckauer:2017}, and cannot 
accommodate other conversion methods, e.g., threshold balancing \cite{Sengupta:2019}. %
We propose a novel
theoretical framework for CNN-to-SNN conversion that covers both weight normalisation \cite{Rueckauer:2017} and threshold balancing \cite{Sengupta:2019} as special cases.
In particular, the proposed framework improves over \cite{Rueckauer:2017} by adopting a thresholding mechanism to quantify the accumulated current into spikes in SNN, and
extends the threshold balancing mechanism to be compatible with batch normalisation and bias. 

We will work with the spiking rate of each SNN neuron $i$ at layer $n$, defined as $r_i^n(t) = N_i^n(t)/t$,
where $N_i^n(t)$ is the number of spikes generates in the first $t$ timesteps by neuron $i$ at layer $n$. We remark that, it is possible that $r_i^n(t)>1$, i.e., multiple spikes in a single timestpes, in which case the latency is increased to process extra spikes. 

Our 
framework is underpinned by Proposition ~\ref{prop:main}.
\begin{proposition} \label{prop:main}
In the CNN-to-SNN conversion, if the first layer CNN activation $a_i^1$ and the first layer SNN current $Z^1_i(t)$ satisfy the following condition
\begin{align}\label{eq:normalisation}
& \frac{1}{T}\sum_{t=1}^{T} Z^1_i(t) = a_i^1,
\end{align}
where $T$ is a predefined maximum timesteps,
then 
 the SNN spiking rate at time step $t$
 can be iteratively computed by
\begin{equation} \label{eq:spiking-rate}
\begin{array}{l}
\displaystyle r^n_{i}(t) = \frac{1}{V^n_{thr}} \Big(\sum^{M^{n-1}}_{j=1}W^n_{ij}r_j^{n-1}(t)+b_i^n \Big)- \Delta_i^n(t)
\end{array}
\end{equation}
with $\Delta_i^n(t) \triangleq \displaystyle {V^n_i(t)}/{(t V^n_{thr})}$
representing the residual spiking rate. Initially, the spiking rate of neuron $i$ at the first layer is $r^1_{i}(t) = a^1_{i}/V^1_{thr} - \Delta_i^1(t) $. 

\end{proposition}

\begin{remark}
The spiking rate in Equation \eqref{eq:spiking-rate} is a generalised form of those using weight normalisation (WN) \cite{Rueckauer:2017} and threshold balancing (TB) \cite{Sengupta:2019}. 
When keeping $\displaystyle V^1_{thr}=1$, by normalising $W^n_{ij}$ we obtain WN; when keeping $W^n_{ij}$ unchanged, by normalising $V^n_{thr}$ we obtain TB. When applying a scaling factor $\alpha^n$ to the threshold $V^n_{thr}$, Proposition \ref{prop:main} recovers \cite{Han:2020}.


\end{remark}

{
The condition in Equation \eqref{eq:normalisation} bridges between the activations in CNNs and the accumulated currents in SNNs, i.e.,  \emph{within the duration of a spike train, the average accumulated current equals to the CNN activation}. This is key to our 
theoretical framework, and different from some previous conversion method such as \cite{Rueckauer:2017},  which bridges between activations and firing rates. 
This \emph{activation-current association} 
is reasonable because it aligns with the intuitions that (i) given a fixed spiking rate, a greater CNN activation requires a greater accumulated current 
in the SNN; and (ii) given a pre-trained CNN, more input spikes lead to increased current in the SNN.
}

Proposition \ref{prop:main} suggests that an explicit, optimised control on the currents may bring benefits to the spiking rate (so as to reduce energy consumption) and the residual current (so as to reduce the accuracy loss) simultaneously. Firstly, a normalisation on the currents $Z_i^n(t)$ is able to control the spike number, with
its  details being given in Section \ref{sec:current_norm}. 
%
Secondly, the error term $\Delta_i^n(t)$
will accumulate in deeper layers, causing lower spiking rate in the output layer~\cite{Rueckauer:2017}. The thresholding technique in Section \ref{sec:thresholdingmechanism} will be able to reduce impact from such error. Thirdly, we need to maintain the consistency between CNN and SNN so that the above control can be effective, as in Section~\ref{sec:batchnormalisation}. 

The input is
encoded into spike train {via Poisson event-generation process \cite{Sengupta:2019} or  interpreting the input as constant currents \cite{Rueckauer:2017}. In this paper we select the latter. } 


\subsection{ECC-Based Conversion Techniques}


We develop three ECC-based techniques, including  current normalisation (CN), thesholding for residual elimination (TRE), and consistency maintenance for batch-normalisation (CMB). 
%
Figure \ref{fig:CNN-SNN-conv} illustrates CN and TRE,
where the $n$-th layer of CNN is on the top and the corresponding conversed SNN layer is at the bottom.
In the converted SNN layer, the sequences of spikes from the previous layer are aggregated, from which the current $Z_i^n(t)$ is accumulated in the neurons, and normalised by a factor (see Equation (\ref{equ:currentnormalisation}) below)
to ensure that {the increase of current at each timestep} is within the range of $[0,1]$.  
The membrane potential $V_i^n(t)$ is produced according to Equation \eqref{eq:em_potential}, followed by a spike generating operation as in Equation \eqref{equ:originaltheta} once $V_i^n(t)$ exceeds the threshold $V_{thr}^n=\CAF_n$. The parameter  $\CAF_n$ is the {c}urrent {a}mplification {f}actor, which will be explained in Section~\ref{sec:current_norm}.
{The residual current $\Delta_i^n$ at the end of spike train indicates the information loss in SNNs.
}





\subsubsection{Current Normalisation (CN)} 
\label{sec:current_norm}
At layer $n$, before spike generation, CN normalises the current $Z^n_i(t)$  
by letting 
\begin{align}\label{equ:currentnormalisation}
    Z^n_i(t) \gets \frac{\lambda_{n-1}}{T \lambda_n} Z^n_i(t), \quad \forall\;t=1,\dots,T
\end{align}
where $\lambda_n \triangleq \max_i \{a_i^n\}$ for $n=1,2,\dots,N$. We have $\lambda_0=1$ when the input has been normalised into $[0,1]$ for every feature.
The benefit of CN is two-fold:
\begin{itemize}
    \item By CN, the maximum number of spikes fed into the SNN is under control, i.e., we can have a direct control on the energy consumption. 
    \item It facilitates the use of a positive integer $V_{thr}^n=\CAF_n$ as the threshold to quantify the current, which is amplified by factor of $\CAF_n$, for spike generation. In doing so, the neuron with maximum current can generate spike at every time step. 
\end{itemize}


We randomly choose $\CAF_n = 100$ for $V_{thr}^n$ and normalise weights for all experiments, except quantised SNN. Since the scalar of quantized weights in each layer will be absorbed into the threshold, we will get a different threshold for each layer. The quantisation process is explained in Section \ref{sec:hyperparameter}.


To achieve CN, the following conversion can be implemented to normalise weights and bias as follows. 
\begin{equation}
W^n_{ij} \gets \CAF_n \frac{\lambda_{n-1}}{\lambda_n} W^n_{ij},~ b^n_i \gets \frac{\CAF_n b^n_i}{\lambda_{n}},~ V^n_{thr} \gets \CAF_n. 
\end{equation}
{Note that, the next layer will amplify the incoming current back to its original scale before its normalisation. When $\CAF_n = 1 ~ or ~ \lambda_{n}/\lambda_{n-1}$, the conversions correspond to weight normalisation in \cite{Rueckauer:2017} and threshold balancing in \cite{Sengupta:2019} respectively.}


\subsubsection{Thresholding for Residual Elimination (TRE)}\label{sec:thresholdingmechanism}

According to Equation \eqref{eq:spiking-rate}, the error increment after conversion is mainly caused by the residual information, $\delta_i^n(T) \in [0,V^n_{thr}]$, which remains with each neuron after $T$ timesteps and cannot be forwarded to higher layers. To mitigate such errors, 
we propose a technique TRE to keep $\delta_i^n(T)$ under a certain value (half of $V_{thr}^n$ as in our experiments).
In particular, we add extra current to each neuron 
in order to have $\eta V_{thr}^n$ increment on each membrane potential, where $\eta \in [0,1)$. Specifically, we update the bias term $b_i^n$ of neuron $i$ at layer $n$ as follows 
\begin{equation}
b_i^n(t):= b_i^n(t)+\eta V^n_{thr}/T
\end{equation}
for every timestep $t$. 
Intuitively, we slightly increase synaptic bias for every neuron at every step, so that a small volume of current is pumped into the system continuously. 

The following proposition says that this TRE technique will be able to achieve a reduction of error range, which directly lead to the improvement to the accuracy loss.  
\begin{proposition}
Applying TRE will lead to
    \begin{align}
        \Theta_i^n(T) = \left\{ \begin{matrix}
       1, & \text{if $V_i^n(T) > (1-\eta) V^n_{thr}$}\\
        0, & \text{otherwise.\hspace{2cm}}
        \end{matrix}
        \right.
    \end{align}
for timestep $T$, as opposed to Equation (\ref{equ:originaltheta}). By achieving this, the possible range of errors is reduced from {$[0,V^n_{thr})$ to $[0,(0.5+|0.5-\eta|) V^n_{thr})$}. 
\end{proposition}
%



We remark that, deploying TRE will increase at most one spike per neuron at the first layer and continue to affect the spiking rate at higher layers. This is the reason why we have slightly more spike operations than \cite{Rueckauer:2017}, as shown in 
Figure S1C in SM.
A typical value of $\eta$ is 0.5.

\subsubsection{{Consistency Maintenance for Batchnormalisation (CMB)}}\label{sec:batchnormalisation}

Batch normalisation (BN) \cite{pmlr-v37-ioffe15} accelerates the convergence of CNN training and improves the generalisation performance. The role of BN is to normalise output of its previous layer, which allows us to add the normalised information to weights and bias in the previous layer. 
We consider a conversion technique CMB to maintain the consistency between SNN and CNN in operating BN layer, by requiring a {c}onstant for {n}umerical {s}tability $\epsilon$, as follows.
\begin{align}
&\hat{W}^n_{ij} = \frac{\gamma_i^n}{\sqrt{\sigma_i^{n2}+\epsilon}} W^n_{ij}\\
&\hat{b}^n_i = \frac{\gamma_i^n}{\sqrt{\sigma_i^{n2}+\epsilon}}(b^n_i - \mu_i^n) + \beta_i^n
\end{align}
where $\gamma_i^n$ and $\beta_i^n$ are two learned parameters, $\mu_i^n$ and $\sigma_i^{n}$ are mean and variance. $\epsilon$ is platform dependent: for Tensorflow it is default as 0.001 and for PyTorch it is 0.00001. 
The conversion method in \cite{Rueckauer:2017} 
does not consider $\epsilon$, and we found through a number of experiments that a certain amount of accuracy loss can be observed consistently.
Figure~\ref{fig:ouringredients} shows the capability of CMB in reducing the accuracy loss.

%

\section{Experiment}\label{sec:experiments}
%
%
%

We implement the ECC method 
and conduct 
an extensive set of experiments to validate it. 
We consider its comparison with the state-of-the-art CNN-to-SNN conversion methods on images and DVS inputs (Section~\ref{sec:TMTScomparison} and Section~\ref{sec:dvs}, respectively), the demonstration of its working with batch-normalisation (Section~\ref{sec:expbn}), its robustness with respect to hardware deployment (Section~\ref{sec:hyperparameter}), and an ablation study (Section~\ref{sec:ablation}). Due to the space limit, we present a subset of the results -- the Supplementary Material (SM) include more experimental results. 
We fix $\kappa_n=100$ and $\epsilon=0.001$ throughout the experiments. 

\begin{figure*}
    \centering
    \includegraphics[width=\textwidth]{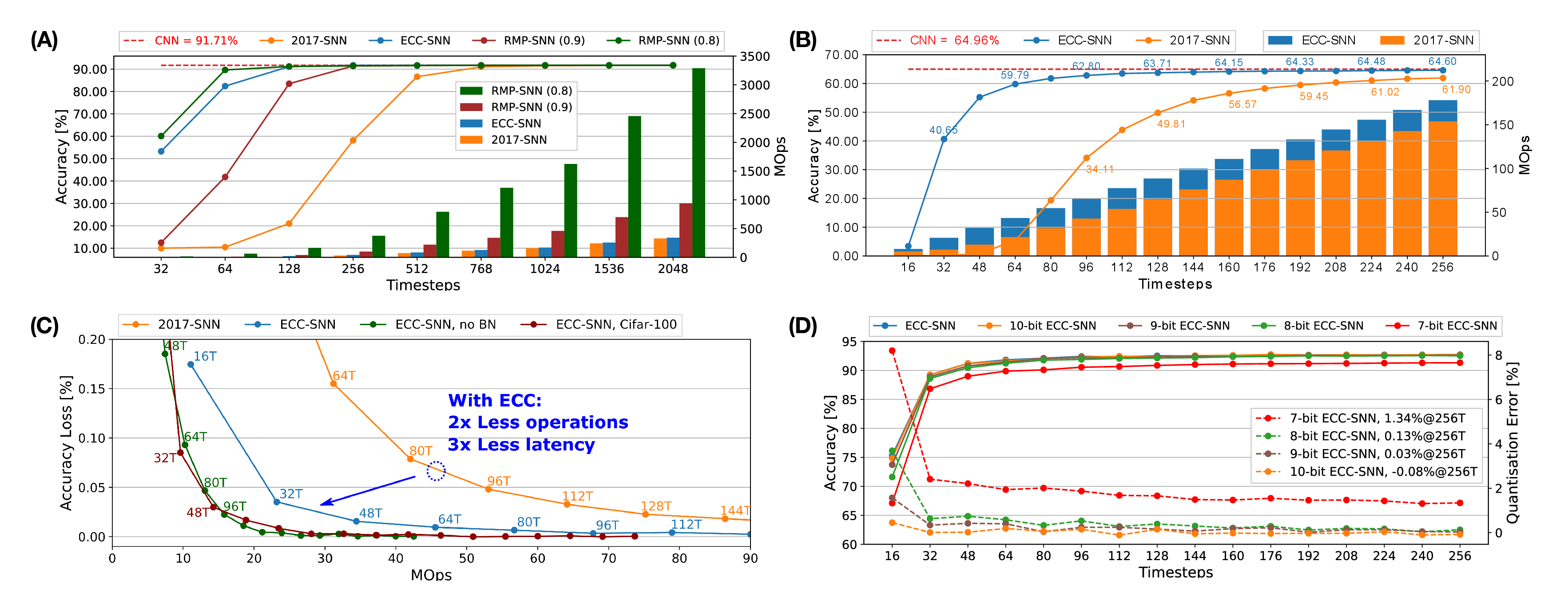}
    \caption{\textbf{(A)}\label{fig:TMTScomparison} Accuracy and energy consumption (MOps) w.r.t. timesteps for CIFAR-10. \textbf{(B)}\label{fig:TMBNIMAGENET} Accuracy and energy consumption (MOps) w.r.t. timesteps, for ImageNet  (Top-1 Acc). \textbf{(C)} \label{fig:TMBNCIFAR10} Accuracy loss and latency w.r.t. energy consumption (MOps),  for CIFAR-10 and CIFAR-100. \textbf{(D)} \label{fig:accuracyerrorbits} Accuracy and quantisation error w.r.t. timesteps, for CIFAR-10. }
    \label{fig:exp}
\end{figure*}

In this section, `2017-SNN' denotes the method proposed in \cite{Rueckauer:2017}.
`RMP-SNN(0.8)' and `RMP-SNN(0.9)' denote the method in \cite{Han:2020}, with different parameter 0.8 or 0.9 as co-efficient to $V_{thr}$. `ECC-SNN' is the our method.
We remark that, it is shown in \cite{Han:2020} that its conversion method outperforms that of  \cite{Sengupta:2019}, so we 
only compare with \cite{Han:2020}. Moreover, 
we may write `Method@$n$T' to represent the specific `Method' when considering the spike trains of length $n$. Note that, only the CNN model in Figure \ref{fig:TMTScomparison}A was trained without bias and BN, in order to have a fair comparison with RMP-SNN techniques. Since BN layers play an important role in training a high performance CNN and has the benefit of lowering the latency (c.f. Section \ref{sec:expbn}),  we believe it is essential to include it in CNN training. Therefore, we do not compare with \cite{Han:2020} (i.e., RMP-SNN) and \cite{Sengupta:2019}  in other experiments because they do not work with BN.

Before proceeding, we explain how to estimate energy consumption.
For CNNs, it is estimated through the multiply–accumulate (MAC) operations
\begin{align}
&\textit{MAC operations for CNNs}: \sum_{n=1}^N (2f^n_{in}+1)M^n
\end{align}
where 
$f^n_{in}$ is the number of input connections of the $n$-th layer. The number of MAC operations are fixed when the architecture of the network is determined. For SNNs, the synaptic operations are counted to estimate the energy consumption of SNNs \cite{Merolla:2014, Rueckauer:2017}, as follows.  
\begin{align}
&\textit{Synaptic operations for SNNs}: \sum_{t=1}^T \sum_{n=1}^N f^n_{out}s^n
\end{align}
where 
$f^n_{out}$ is the number of output connections and $s^n$ is the average number of spikes per neuron, 
of the $n$-th layer.

\subsection{Experimental Settings} 

We work with both image datasets (CIFAR-10/100 \cite{cifar10:2007} and ImageNet \cite{ILSVRC15}) and DVS datasets (CIFAR-10-DVS \cite{li2017cifar10dvs}) on several architectures \cite{VGG:2014} (VGG-16, VGG-19, and VGG-7).
All the experiments are conducted on a CentOS Linux machine with
two 2080Ti GPUs and 11 GB memory. 

\subsection{Comparisons with State-of-the-Art}\label{sec:TMTScomparison}

Figure~\ref{fig:TMTScomparison}A presents a comparison between 2017-SNN, RMP-SNN, and ECC-SNN on both accuracy and energy consumption w.r.t. the timesteps, on VGG-16 and CIFAR-10. We note that, both RMP-SNN and ECC-SNN outperform 
2017-SNN, in terms of the number of timesteps to reach near-zero accuracy loss. 
Furthermore, ECC-SNN is better than RMP-SNN(0.9) and competitive with RMP-SNN(0.8) in terms of reaching near-zero accuracy loss under certain latency. Specifically, both ECC-SNN and RMP-SNN(0.8) require 128 timesteps and RMP-SNN(0.9) requires 256 timesteps. Importantly, we note that, both RMP-SNN(0.8) and RMP-SNN(0.9) consume much more energy, measured with MOps, than ECC-SNN. Actually, ECC-SNN does not consume significantly more energy than 2017-SNN.   
Similar results can be extended to large dataset such as ImageNet. 
Moreover, to investigate further into the energy consumption, 
Figure \ref{fig:TMBNIMAGENET}B presents a comparison with 2017-SNN. All the above results show that ECC-SNN significantly reduces the latency, easily reaches the near-zero loss, and costs a minor increase on the energy.  

The above results, together with those in SM 
(Figure S1A, Figure S1C, Figure S1D, and Figure S1E),
reflect exactly the advantage of using ECC-SNN. That is, RMP-SNN(0.8) and ECC-SNN are the best in achieving near-zero accuracy loss with low latency, but RMP-SNN requires significantly more energy than the other two methods. Therefore, \emph{ECC-SNN achieves the best when considering energy, latency, and accuracy loss}. 

Batch-normalisation (BN) has become indispensable to train CNNs, so we believe a CNN-to-SNN method should be able to work with it. After demonstrating clear advantage over RMP-SNN, for the rest of this section, we will focus on the comparison with 2017-SNN, which deals with BN. 
We trained CNNs using Tensorflow by having a batch-normalisation layer after each convolutional layer. 

\subsection{Batch-normalisation (BN)}\label{sec:expbn}

Figure~\ref{fig:TMBNCIFAR10}C considers the impact of working with BN.
Comparing with 2017-SNN, ECC-SNN achieves similar accuracy loss by taking 2x less MOps and 3x less latency. 
Moreover, to achieve the same accuracy loss, ECC-SNN without BN, i.e., ECC applies on CNNs without BN layers,  requires significantly more timesteps, with slightly less MOps. 
%
Moreover, our other experiments show that,  
RMP-SNN (0.8), without BN in its method,  can only achieve 48.32\% in 256T. With BN, 2017-SNN can achieve 49.81\% in 128T. ECC-SNN further improves on this, achieving 63.71\% in 128T. That is, \emph{batch-normalisation under ECC-SNN can help reduce the latency}.  
This is somewhat surprising, and we believe further research is needed 
to investigate the formal link between BN and latency.



\subsection{Robustness to Quantisation}\label{sec:hyperparameter}


%
Figure~\ref{fig:accuracyerrorbits}D (and 
Figure S1B
in SM) present how the change on the number of bits to represent weights may affect the accuracy and the quantisation error. This is an important issue, as the SNNs will 
be deployed on neuromorphic chip, such as Loihi \cite{Loihi:2018} and TrueNorth
\cite{TrueNorth:2015}, or FPGA, which may have different configurations. For example, Loihi can have weight precision at 1-9 bits. Floating-point data, both weights and threshold, can be simply converted into fixed-point data {after CN} in {two steps: normalising the weights into range [-1,1] and scaling the threshold using the same normalisation factor, and then} multiplied with $2^b$, where $b$ is the bit width \cite{Ju:2019,Sze:2017}. From Figure~\ref{fig:accuracyerrorbits}D and 
Figure S1B
, the reduction from 32-bit to 10-, 9-, 8- and 7-bit signed weights does lead to drop on the accuracy, but unless it goes to 7-bit, the accuracy loss is negligible. This shows that, \emph{our ECC method is robust to  hardware deployments}.


\subsection{DVS Dataset}\label{sec:dvs}

\begin{figure*}[h!]
	\begin{minipage}{0.48\textwidth}
    \centering
    \includegraphics[width=\textwidth]{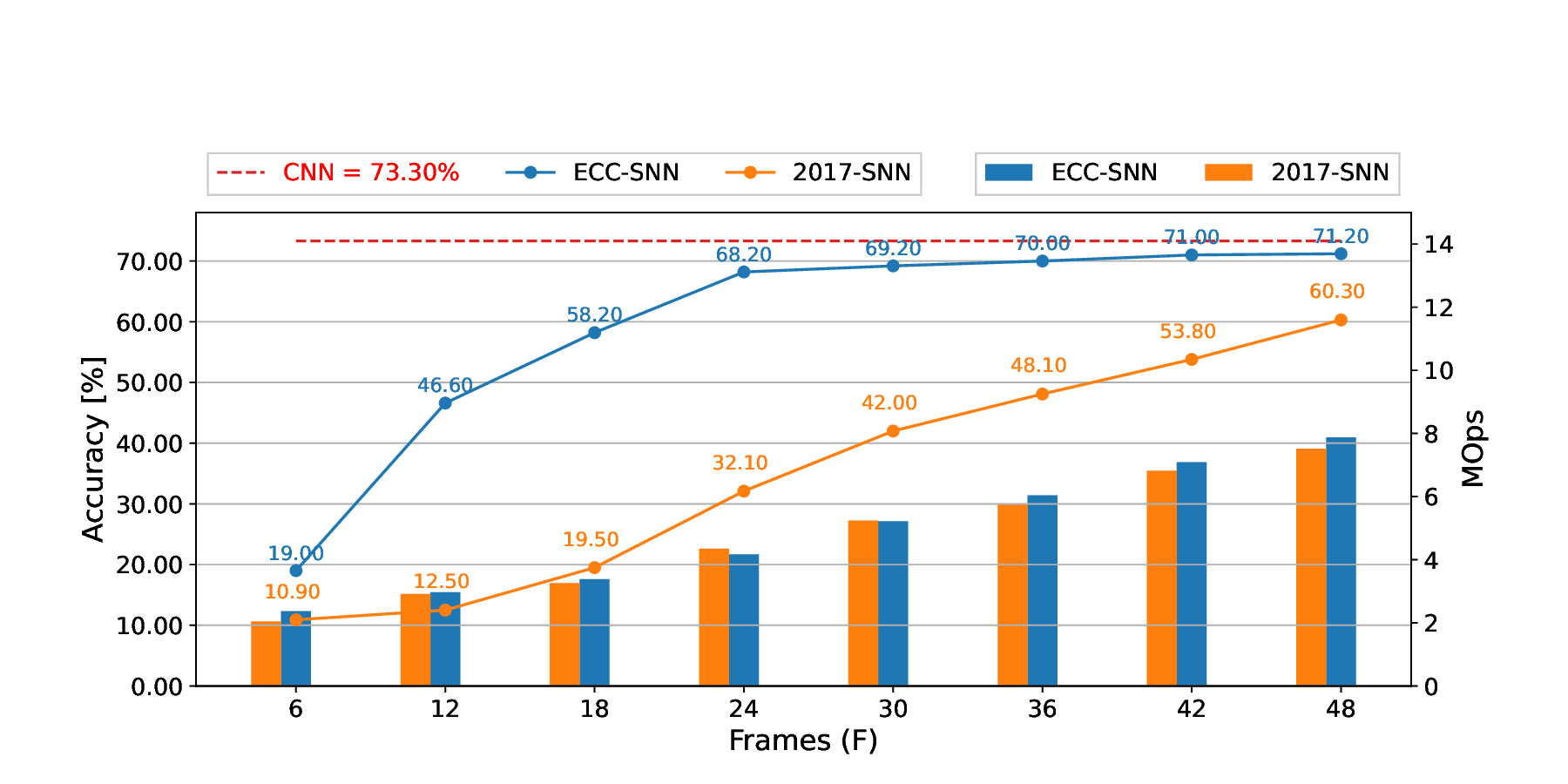}
    \caption{Accuracy and energy consumption (MOps) w.r.t. frames, between 2017-SNN and ECC-SNN, for CIFAR-10-DVS and VGG-7.}
    \label{fig:dvsvgg7}
    \end{minipage}
	\begin{minipage}{0.48\textwidth}
	\centering
	\vspace{20pt}
	\includegraphics[width=\textwidth]{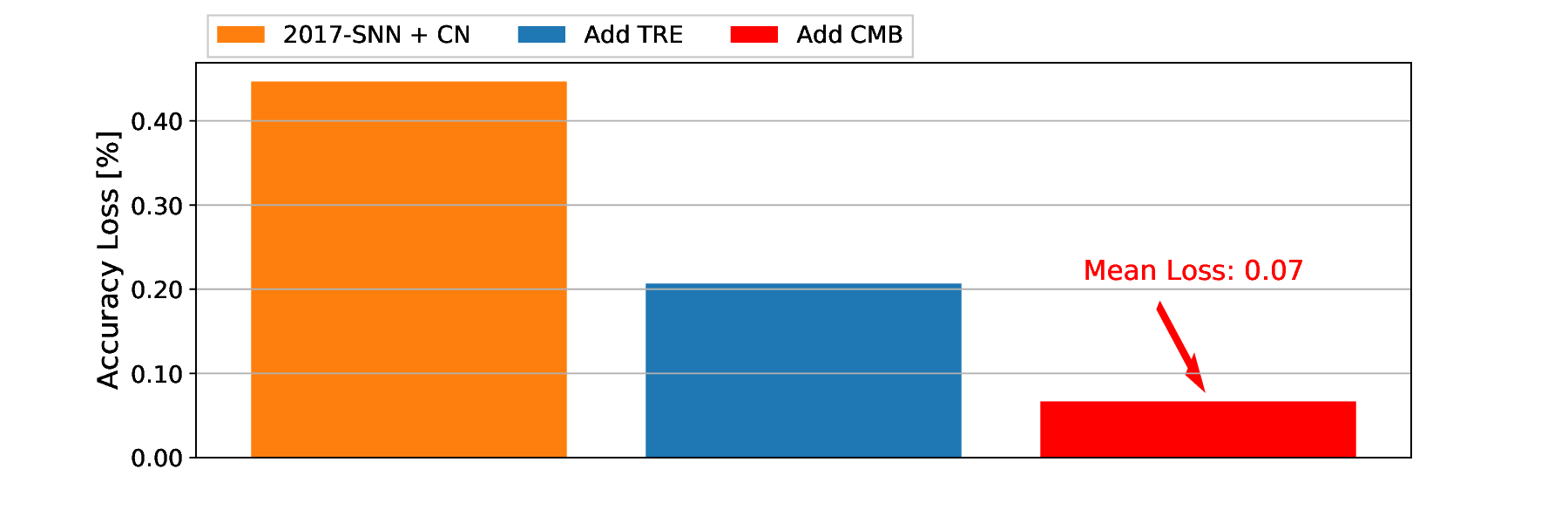}
	\vspace{5pt}
	\caption{Contribution of CN, CMB, and TRE to the reduction of {mean accuracy loss}, for CIFAR-10 and VGG-16.}
	\label{fig:ouringredients}
\end{minipage}
\end{figure*}

CIFAR-10-DVS \cite{li2017cifar10dvs} is a benchmark dataset of DVS inputs, consisting of 10,000 inputs 
extracted from CIFAR-10 dataset using a DVS128 sensor. The resolution of data is 128x128. We preprocess the data following \cite{wu2021training, kugele2020efficient}, select the first 1.3s of the event stream, and down-scale the input into 42x42. For each dimension of an input, we calculate the number of spikes over the 1.3s simulation and normalise with a constant representing the maximum number of spikes. 
During SNN processing, as shown in Figure \ref{fig:introductionDescription}(B), the latency is based on the spikes. 
The experiments using VGG7 (Figure \ref{fig:dvsvgg7}) , VGG16 and ResNet-18 \cite{Sengupta:2019} (Figure 
S2A
and 
S2B in SM) show that ECC-SNN performs much better than 2017-SNN, who also work with BN layer. Moreover,  in Table \ref{tab:dvscomparison}, we compare with a few methods including a recent direct training method. We can see that ECC-SNN can always achieve better accuracy, less frames, and less energy consumption. 

Moreover, we recall the results shown in Figure~\ref{fig:introductionDescription} concerning the comparison between DVS inputs and images. For the same problem, if we choose the  deployment workflow of ``training a CNN $\rightarrow$ converting into an SNN $\rightarrow$ deploying on edge devices with e.g., event camera'', we may consume 10+ times less energy (7.52MOps vs 90MOps for CIFAR-10) by taking DVS inputs. Both are in stark contrast with the other deployment workflow ``training a CNN $\rightarrow$ deploying on edge devices with camera'', which costs much more energy (307MOPs and 657MOps, respectively).


\begin{table}[h!]
    \centering
    \begin{tabular}{| l | l| l | l | l | }
    \hline
    Method  & Accuracy & N$^*_{f}$ & MOps \\
    \hline   
    Direct training (VGG7) \cite{wu2021training} & 62.50 & - & - \\
    2017-SNN (DenseNet) \cite{kugele2020efficient} & 65.61 & 60 & 1,551      \\
    ECC-SNN (VGG16) & 71.20 & \textbf{48} & 66.79 \\
    ECC-SNN (VGG7) & \textbf{71.30} & \textbf{48}  & \textbf{7.52}  \\
    \hline
    \end{tabular}
    \raggedright
    *N$_{f}$ is the number of frames.
    \caption{Comparison of SNN accuracy, latency and energy consumption (MOps), between direct training, 2017-SNN and ECC-SNN, for Cifar-10-DVS}
    \label{tab:dvscomparison}
\end{table}


\subsection{Ablation Study}\label{sec:ablation}

To understand the contributions of the three ingredients of ECC-SNN,
i.e., CN, CMB, and TRE, 
we conduct an experiment on VGG-16 and CIFAR-10, by gradually including technical ingredients to see their respective impact on the accuracy loss. Figure~\ref{fig:ouringredients} shows the histograms of the mean accuracy losses in 256T, over the 281-283th epochs. 
We see that, every ingredient plays a role in reducing the accuracy loss, with the TRE and CN being lightly 
Moreover, 
We also consider the impact of $\eta$ (as in Figure
S1F
of SM).




\section{Conversion Optimised through Distribution-Aware CNN Training}\label{sec:clippedmethods}

Up to now, all methods we discussed and compared with, including our ECC method, are focused on optimising the CNN-to-SNN conversion, without considering whether or not the CNN itself may also play a role in eventually obtaining a good-performing SNN. In this section, we will discuss several recent techniques that include the consideration of CNN training, and show that our ECC method can also improve them by optimising the CNN-to-SNN conversion.   

As noted in Section~\ref{sec:current_norm} that the maximum value of activations is a key parameter in the conversion. Based on this, \cite{Rueckauer:2017, lu2020exploring} suggest that a particular percentile from the histogram on the CNN activation may improve conversion efficiency. One step further, \cite{yu2020low} suggests that a good distribution with less outliers on CNN activation can be useful for 
quantisation.
Therefore, we call these techniques distribution-aware CNN training techniques, to emphasise that they are mainly focused on optimising the CNN training through enforcing good distributions on the activations. 

\begin{figure*}[h!]
	\includegraphics[width=\textwidth,height=0.48\textwidth]{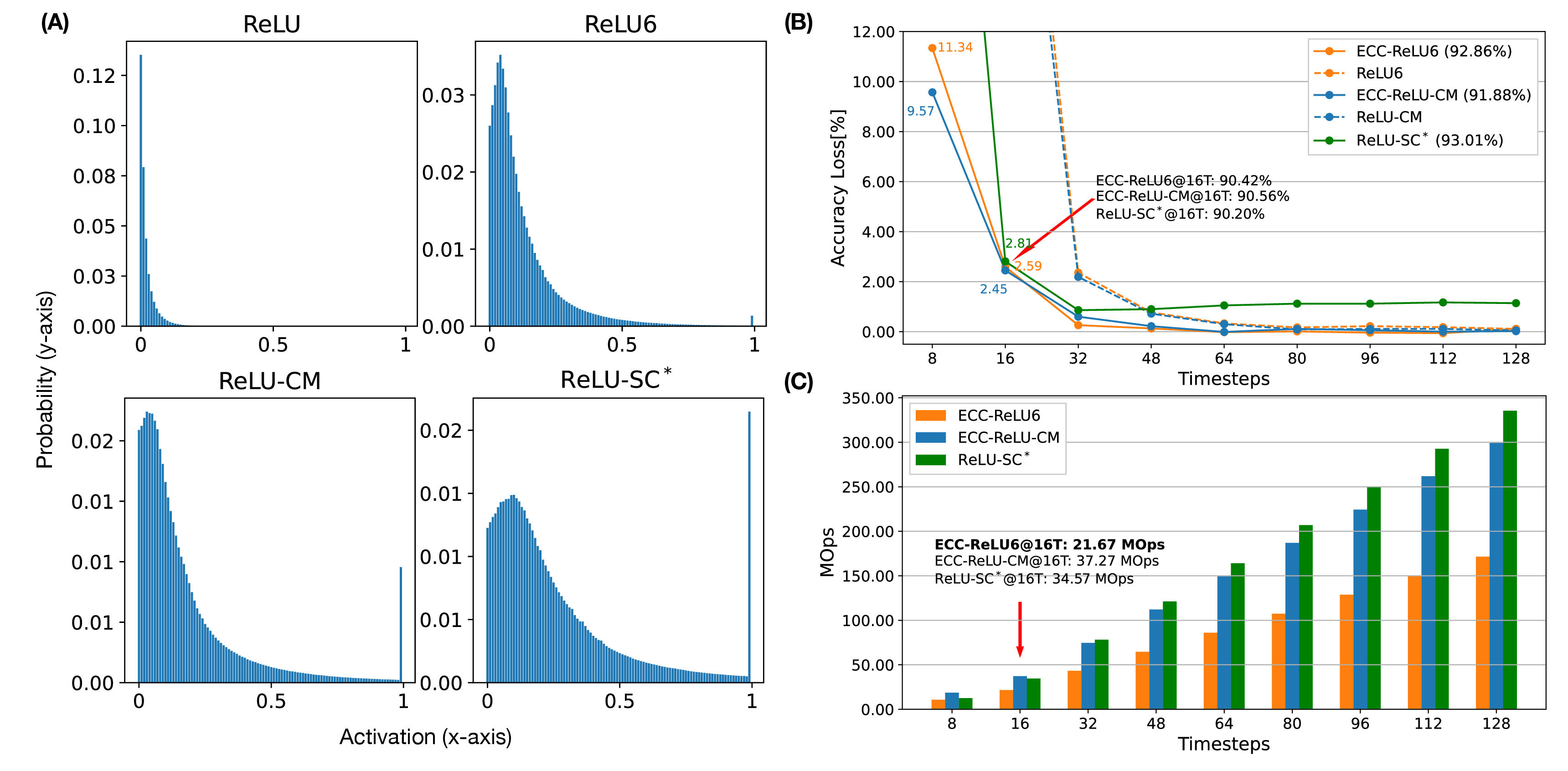}
	\caption{Comparison of SNNs using different clipping methods, ReLU6 (\cite{lin2019defensive, jacob2018quantization}), ReLU-CM (\cite{yu2020low}), ReLU-SC* (\cite{deng2021optimal}), for CIFAR-10 on VGG-16. \textbf{(A)} Normalised activation distribution of the first layer. \textbf{(B)} Accuracy w.r.t. timesteps. \textbf{(C)} Energy consumption (MOps) w.r.t. timesteps. *We use ReLU-SC to train an SNN with a fixed timestep (16T), as it does not need extra training.} 
	\label{fig:clippedmethods}
\end{figure*}

To show that our ECC method is complementary to the distribution-aware CNN training techniques, we implement some existing CNN training techniques that can affect activation distribution, and show that ECC can also work with them to achieve optimised conversion. 
Specifically, \cite{yu2020low} notice that the clipped ReLU can enforce small activation values (i.e., close to zero) to become greater, and eventually reshape the distribution from a Gaussian-like distribution to a uniform distribution. 
We follow this observation to train CNN models with different clipping methods,
including ReLU6 (clipped by 6) from \cite{lin2019defensive, jacob2018quantization}, ReLU-CM (clipped by k-mean) from \cite{yu2020low}, and ReLU-SC (shift and clipped) from \cite{deng2021optimal}.
%
Figure \ref{fig:clippedmethods}A shows that all clipping methods can shift the original activation value (as in the top left figure) to values closer to the mean value (i.e., near the peak area). Such a shift of maximum activation value can significantly reduce the possibility for the maximum value to become an outlier. 

Figure \ref{fig:clippedmethods}B presents a comparison between SNNs obtained through  ReLU6, ReLU-CM and ReLU-SC, with and without the application of ECC. First of all, distribution-aware training can improve the performance. E.g., with clipping methods, the accuracy loss is less than 2.2\%@32T, which is better than ECC-SNN 
(3.2\%@32T) as in Figure \ref{S-fig:TM2017comparison}. Then, we can see that, with ECC, ReLU6 and ReLU-CM can achieve 2x less latency with little performance degradation. Although ECC-ReLU6, ECC-ReLU-CM and ReLU-SC achieve similar accuracy (90.20\% to 90.56\%@16T), ECC-ReLU-CM has the best adaptability to different timesteps. By contrast, ECC-ReLU6 uses 1.5 to 1.7x less operations  at 16T than ReLU-SC and ECC-ReLU-CM, as shown in Figure \ref{fig:clippedmethods}C. 
We only apply ECC to ReLU6 and ReLU-CM, as 
ReLU-SC in \cite{deng2021optimal} is not designed to be adaptable to different timesteps. 

%
The above results show that distribution-aware CNN training and our ECC method can both improve the CNN-to-SNN conversion. While distribution-aware CNN training can 
reduce the accuracy loss, the application of ECC method can further improve the performance of the resulting SNN model. Furthermore, it is worth mentioning that ECC can take the advantage of the accumulated bias current to optimise a single SNN model with respect to different timesteps.

%
%
%
%
%

\section{Discussion}


\textbf{Variants to the Unifying Framework} The current unifying framework (Section \ref{sec:TBWN})
 considers ReLU activation function, which exhibits a linear relation between accumulated current and spiking rate. There are other -- arguably more natural --  features in biological neuron, such as leak, refractory time and adaptive threshold, as discussed in  \cite{kobayashi2009made}. If considering these features, the relation between accumulated current and spiking rate will  become non-linear. To deal with them, it can be an interesting future work to 
 consider extending the unifying framework to address the connection between nonlinear activation functions (e.g., sigmoid) on CNN and the  dynamic properties on SNN.

\noindent \textbf{Hyper-parameters in ECC} Most of the hyper-parameters in ECC-SNN are determined with reasons, such as $\kappa_n$ (Section \ref{sec:current_norm}) and $\eta$ (Section \ref{sec:thresholdingmechanism}),  
while timesteps (T) is determined by practical application according to e.g., required accuracy. Although some gradient-based optimisation methods, such as \cite{Rathi2020Enabling,rathi2021diet,pmlr-v139-li21d}, can improve the SNN to a fixed timestep, ECC allows SNN to be adaptive to different timesteps. In the future, we will consider hyper-parameters tuning methods, as in e.g.,  \cite{parsa2020bayesian}, to further improve ECC-SNN while maintaining its adaptability. 

\section{Conclusion}
We develop a unifying theoretical framework to analyse the conversion from CNNs to SNNs and 
 a new conversion method ECC 
 to explicitly control the currents, so as to optimise accuracy loss, energy efficiency, and latency simultaneously. 
%
By comparing with state-of-the-art methods, we confirm the superior performance of our method. Moreover, we  study the impact of batch-normalisation, and show the robustness of ECC over quantisation. \section*{Funding}
DW is supported by the University of Liverpool and China Scholarship Council Awards (Grant No. 201908320488). \includegraphics[height=8pt]{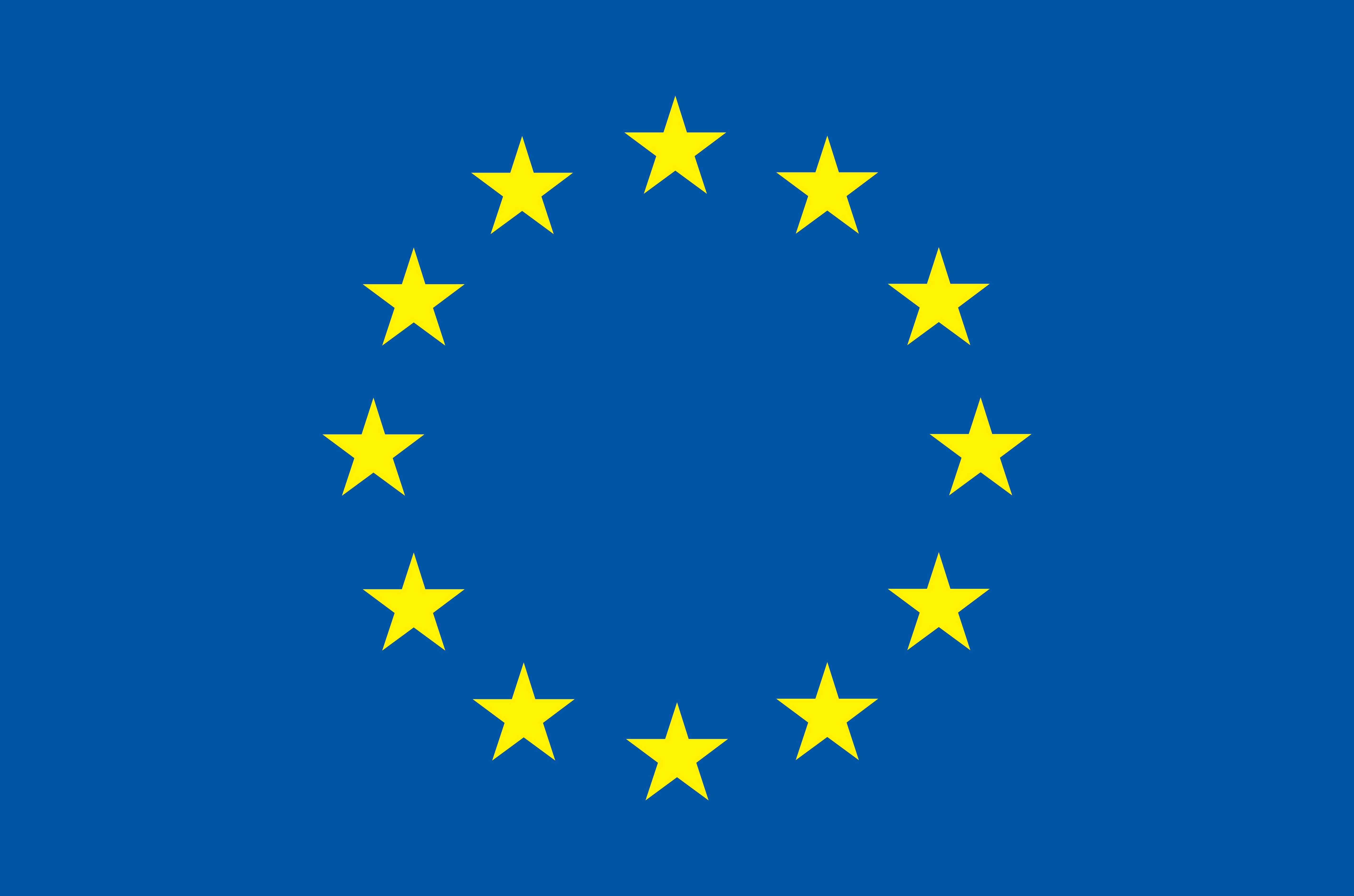} This project has received funding from the European Union’s Horizon 2020 research and innovation programme under grant agreement No 956123. It is also supported by the U.K. EPSRC (through End-to-End Conceptual Guarding of Neural Architectures [EP/T026995/1]).




{\small
\bibliographystyle{ieee_fullname}
\bibliography{egbib}
}

\end{document}


\newcommand{\xiaowei}[1]{{\color{blue}#1}}
\newcommand{\dengyu}[1]{{\color{red}#1}}
\newcommand{\xy}[1]{{\color{cyan}#1}}

\newcommand{\commentout}[1]{}
\newcommand{\CAF}{\kappa}
\newcommand{\TSF}{\alpha}

\newtheorem{proposition}{Proposition}
\newtheorem{remark}{Remark}
\title{The Conversion from Convolutional to Spiking Neural Networks Needs to be Multi-Objective: Accuracy, Latency, and Energy Efficiency}

\author{Dengyu Wu, Xinping Yi, and Xiaowei Huang\\
University of Liverpool, UK\\
{\tt\small \{dengyu.wu,xinping.yi,xiaowei.huang\}@liverpool.ac.uk } 
}

\maketitle
\ificcvfinal\thispagestyle{empty}\fi

The supplementary material of this submission includes this document and a software package SpKeras. This document contains technical proofs for Propisiton 1 and Proposition 2, as well as some additional experiments. The software package includes source codes and its documentation (software framework, README, etc), as well as an example for VGG16 network on CIFAR-10 dataset. 
\section{Proof of Proposition 1}
According to Equation (2), the accumulated current of $i$-th neuron at layer $n$ over the simulation time $T$ can be calculated as
\begin{equation}
\sum_{t=1}^TV^n_i(t) = \sum_{t=1}^T(V^n_i(t-1) + Z^n_i(t) - \Theta_i^n (t) V^n_{thr}).\tag{A1}\label{eq:A1}
\end{equation}

\noindent By using $N_i^n(T) = \sum_{t=1}^T\Theta_i^n (t)$, we obtain
\begin{equation}
N_i^n(T) V^n_{thr}
 = \sum_{t=1}^TZ^n_i(t) - (\sum_{t=1}^TV^n_i(t) - \sum_{t=1}^TV^n_i(t-1))\tag{A2-a}\label{eq:A2-a}
\end{equation}
which further yields
\begin{equation}
\frac{N_i^n(T)}{T} 
 = \frac{1}{T V^n_{thr}}\sum_{t=1}^TZ^n_i(t) - \frac{1}{T V^n_{thr}}(V_i^n(T) - V_i^n(0)).\tag{A2-b}\label{eq:A2-b}
\end{equation}
\noindent According to Equation (3), the current accumulated in hidden layers can be calculated as
\begin{align}
    \sum_{t=1}^T Z^n_i(t) = \sum_{j=1}^{M^{n-1}} W_{ij}^n N_j^{n-1} (T)  + Tb^n_i. \tag{A3}\label{eq:A3}
\end{align}
\noindent By inserting Equation (\ref{eq:A3}) into Equation (\ref{eq:A2-b}) $(n > 1)$ and setting $V(0) = 0$, the spiking rate, $r_i^n(T) = N_i^n(T)/T$, can be iteratively computed by
\begin{equation} 
\begin{array}{l}
\displaystyle r^n_{i}(T) = \frac{1}{V^n_{thr}} \Big(\sum^{M^{n-1}}_{j=1}W^n_{ij}r_j^{n-1}(T)+b_i^n \Big)- \frac{V^n_i(T)}{T V^n_{thr}}. \tag{A4}\label{eq:A4}
\end{array}
\end{equation}
\noindent 
Given the condition in Equation (5), the iterative spiking rate computation can be initialised at first layer $(n = 1)$ as
\begin{equation}
r^1_i(T) = \frac{a^1_i}{V^1 _{thr}} - \frac{V_i^1(T)}{TV^1 _{thr}} \tag{A5}\label{eq:A5}
\end{equation}
\noindent In general, for any $t\in [1,\dots,T]$, the accumulated spiking rate at $t$ can be induced as
\begin{equation} 
\begin{array}{l}
\displaystyle r^n_{i}(t) = \frac{1}{V^n_{thr}} \Big(\sum^{M^{n-1}}_{j=1}W^n_{ij}r_j^{n-1}(t)+b_i^n \Big)- \frac{V^n_i(t)}{t V^n_{thr}} \tag{A6}\label{eq:A6}
\end{array}
\end{equation}
\noindent 
where the spiking rate of the first layer can be initialised as
\begin{equation}
r_i^1(t)  = \frac{a^1_i}{V^1 _{thr}} - \frac{V(t)}{tV^1 _{thr}}. \tag{A7}\label{eq:A7}
\end{equation}
This completes the proof of Proposition 1.

\section{Proof of Proposition 2}

Adding the updated bias term of Equation (9) into Equation \eqref{eq:A4}, we have
\begin{align}
 r^n_{i}(T) &= \frac{1}{V^n_{thr}} \Big(\sum^{M^{n-1}}_{j=1}W^n_{ij}r_j^{n-1}(T)+b_i^n \Big) \notag 
\\ & \qquad \qquad -  \frac{  V^n_i(T) - (1- \eta) V^n_{thr}}{T V^n_{thr}} \tag{B1}\label{eq:B1}
\end{align}
As such, the residual spiking rate due to the residual current turns out to be
\begin{align}
    \Delta_i^n(T) = \frac{  V^n_i(T) - (1 - \eta) V^n_{thr}}{T V^n_{thr}}.  \tag{B2}\label{eq:B2}
\end{align}

If $\Delta_i^n(T) \geq 0$, an extra spike can be generated. This yields the thresholding rule in Proposition 2. 

Moreover, we note that, with thresholding and $\eta$, the residual current will be less than $(0.5+|0.5-\eta|)V_{thr}^n$. Actually, any neuron with more than $(0.5+|0.5-\eta|)V_{thr}^n$ current will automatically incur a spike and lead to a reduced residual current within $[0,(0.5+|0.5-\eta|)V_{thr}^n)$. Together with the previous argument for thresholding rule, this completes the proof of Proposition 2. 


Intuitively, Proposition 2 can be explained as follows. The extra current is injected into the SNN system -- in the form of increased bias term -- to raise the level of residual current beyond the predetermined threshold $V_{thr}^n$. Both the increased residual current and the $\eta$ will take effect in increasing the possibility of generating one extra spike. This is equivalent to decrease the effective threshold at the final timestep $T$ for the residual current, as  in Equation (10). This process ensures that 
the error caused by the residual current can be reduced, because technically the range of residual current retained in the neurons is reduced from $[0,V_{thr}^n)$ to $[0,0.5+|0.5-\eta|V_{thr}^n)$. 


\section{More Experimental Results Comparing to State-of-the-art}

Figure~\ref{fig:TMperformance} shows the accuracy impact of small timesteps, for both ECC-SNN and 2017-SNN. We plot the accuracy of 10 epochs when the CNN training is almost converged, i.e., when the accuracy does not fluctuate significantly with the advance of epochs. 
For 2017-SNN, a smaller timestep may lead to significant performance drop. For example, the accuracy of 2017-SNN@256T is around 92.5\%, while  2017-SNN@128T is around 91.5\%, a 1\% accuracy gap. On the other hand, ECC-SNN does not experience significant accuracy drop. \emph{This observation supports our view that 
we may not need a significant amount of energy (here, the timesteps) to reach an acceptable level of accuracy loss}. 

The last point is further exhibited in Figure~\ref{fig:TM2017comparison}, where we compare only ECC-SNN with 2017-SNN. We can see that, only a minor energy consumption increase from 2017-SNN to ECC-SNN, but ECC-SNN can achieve near-zero accuracy loss with much smaller timesteps. 
\setcounter{figure}{5}
\begin{figure*}[h!]
    \begin{subfigure}{0.48\textwidth}
    \includegraphics[width=\textwidth,height=3cm]{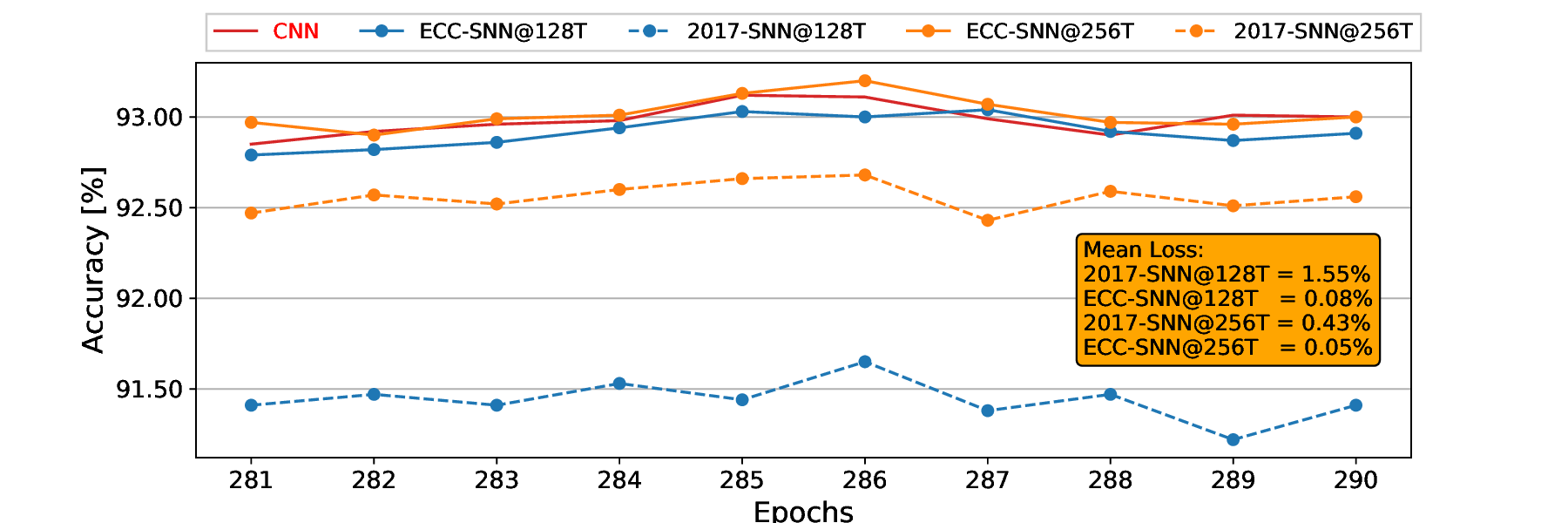}
    \caption{Accuracy w.r.t. training epochs, on SNNs (2017-SNN vs\\ ECC-SNN), for CIFAR-10 (average loss in yellow box)}
    \label{fig:TMperformance}
    \end{subfigure}
    \begin{subfigure}{0.48\textwidth}
    \includegraphics[width=\textwidth,height=3cm]{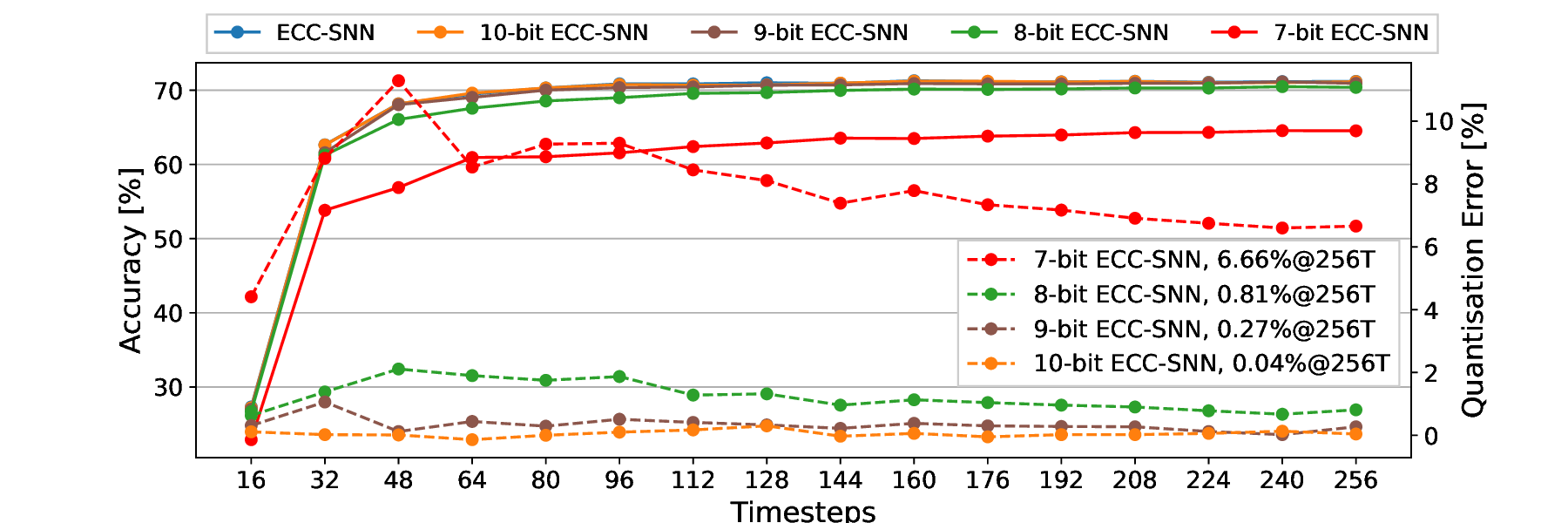}
    \caption{Accuracy and quantisation error w.r.t. timesteps, \\ for CIFAR-100}
    \label{fig:accuracyerrorbits-cifar100}
    \end{subfigure}
    \begin{subfigure}{\textwidth}
    \includegraphics[width=0.48\textwidth,height=3cm]{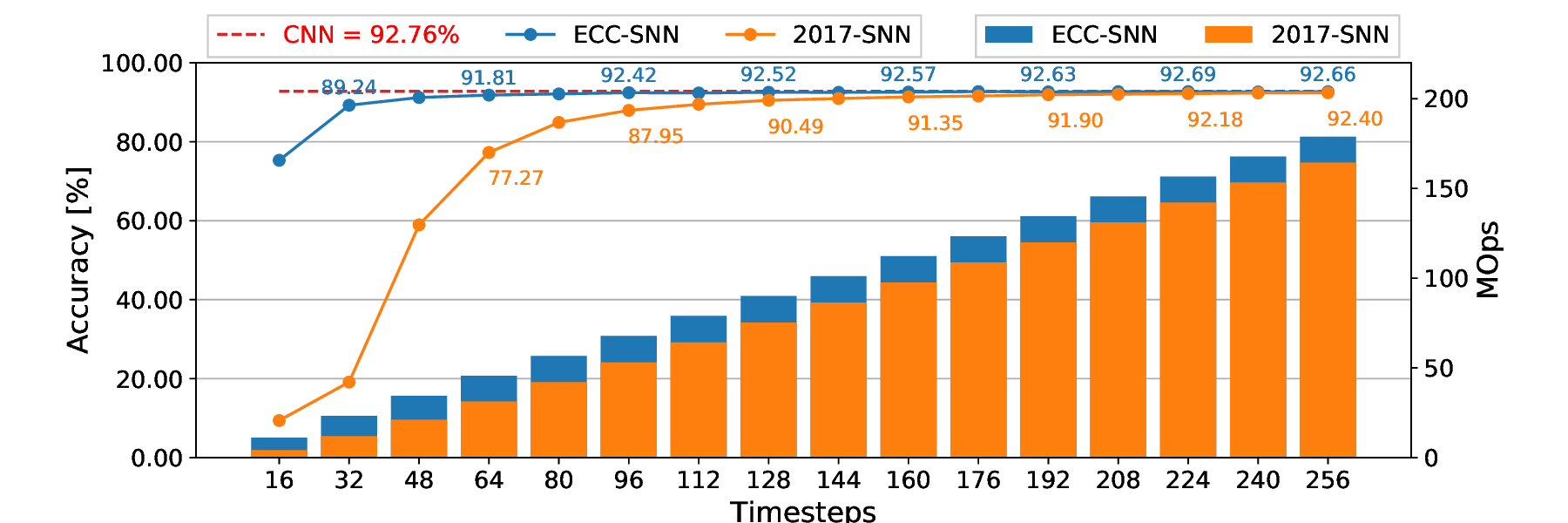}
    \includegraphics[width=0.48\textwidth,height=3cm]{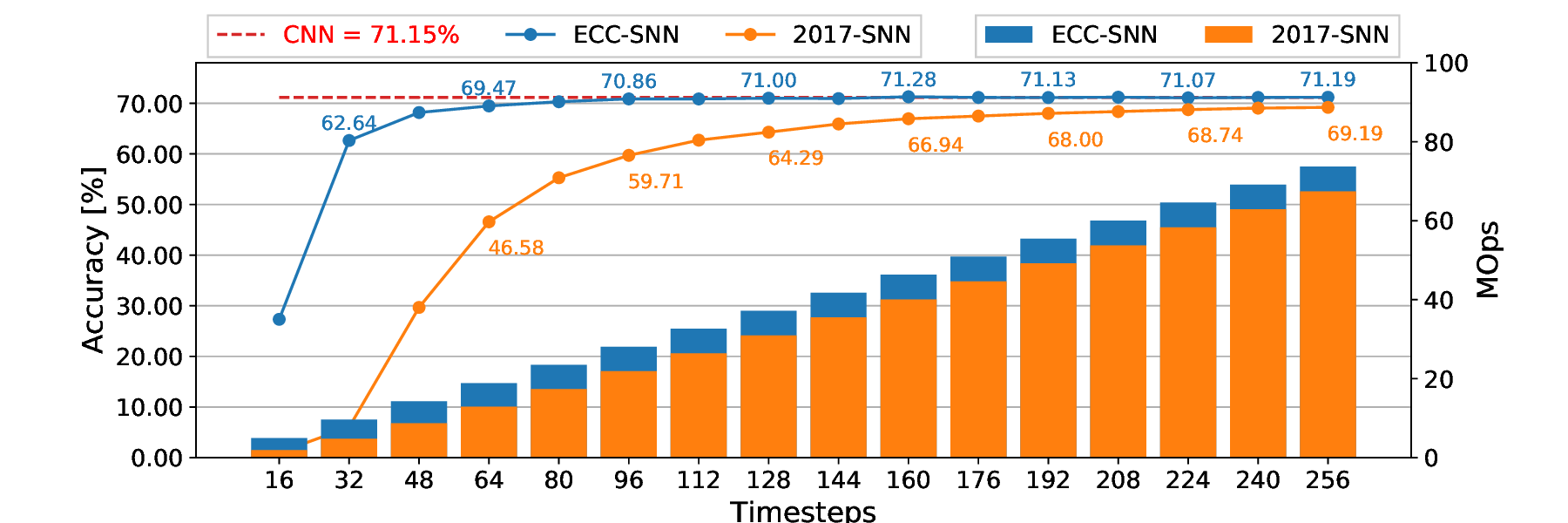}
    \caption{VGG-16: Accuracy and energy consumption (MOps) w.r.t. timesteps, between 2017-SNN and ECC-SNN, for CIFAR-10 (left) and CIFAR-100 (right)}
    \label{fig:TM2017comparison}
    \end{subfigure}    
    \centering
    \begin{subfigure}{0.47\textwidth}
    \resizebox{\columnwidth}{!}{
    \includegraphics{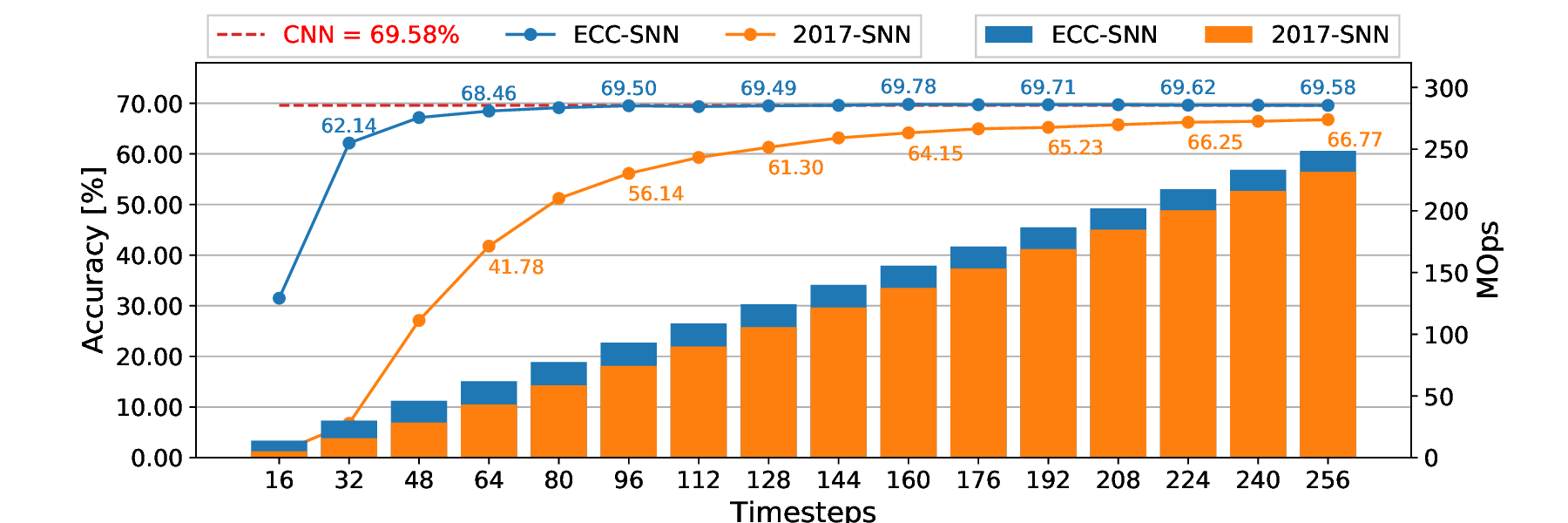}}
    \caption{Accuracy and energy consumption (MOps) w.r.t. timesteps, \\ for CIFAR-100 on VGG-19}
    \label{fig:vgg19a}
    \end{subfigure}
    \begin{subfigure}{0.47\textwidth}
    \resizebox{\columnwidth}{!}{
    \includegraphics{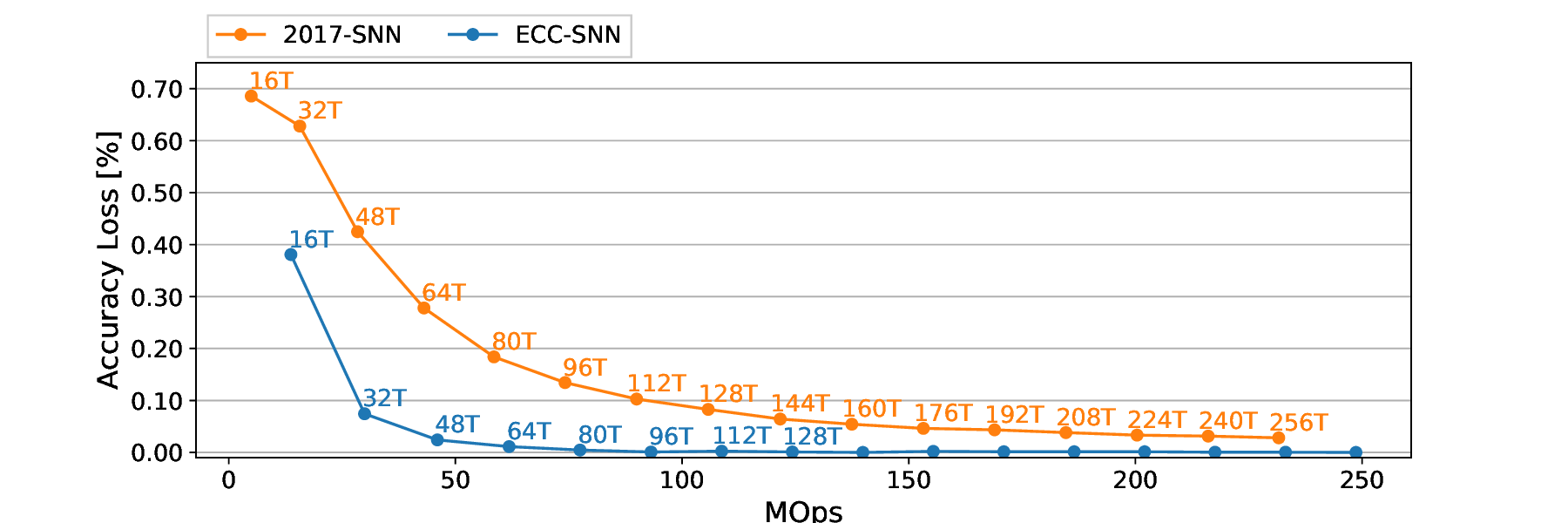}}
    \caption{Accuracy loss and latency w.r.t. energy consumption (MOps), \\ for CIFAR-100 on VGG-19}
    \label{fig:vgg19b}
    \end{subfigure}
    \begin{subfigure}{\textwidth}
    \includegraphics[width=0.48\textwidth,height=2.5cm]{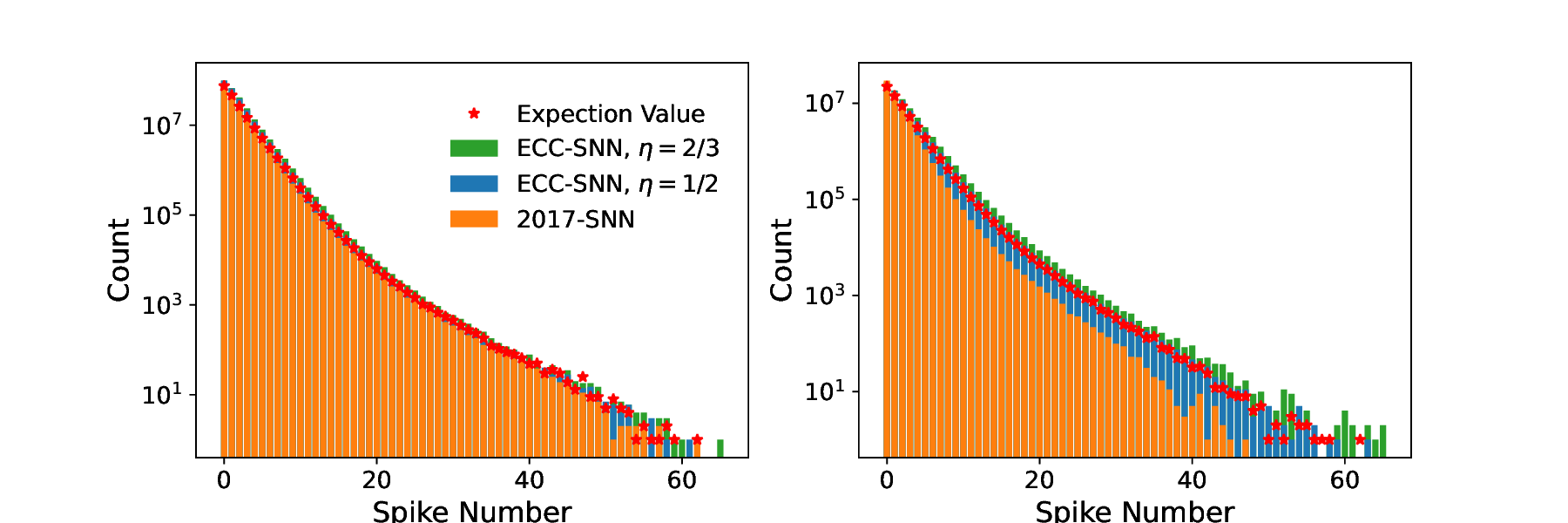}
    \includegraphics[width=0.48\textwidth,height=2.5cm]{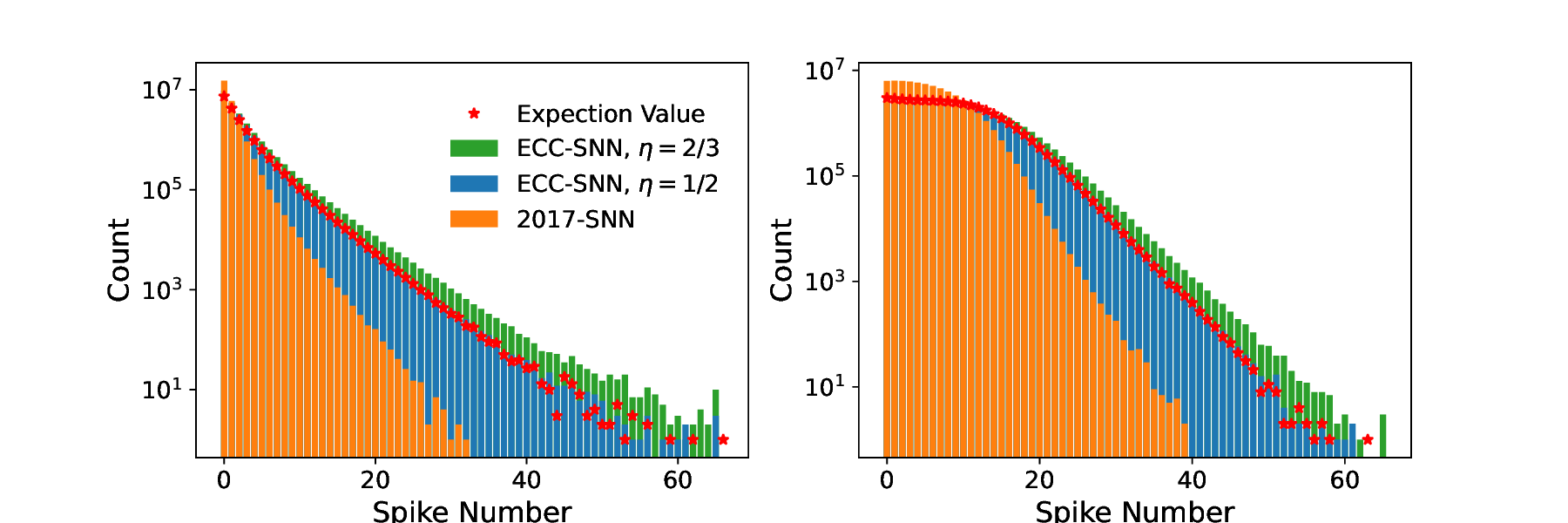}
    \caption{Spike count on 4th, 8th, 12th and 16th layer (from left to right) with 64 timesteps for CIFAR-100}
    \label{fig:spikecount}
    \end{subfigure}
    \caption{Results for Cifar-10/100}
\end{figure*}

\begin{figure*}[h!]
    \begin{subfigure}{0.48\textwidth}
    \includegraphics[width=\textwidth]{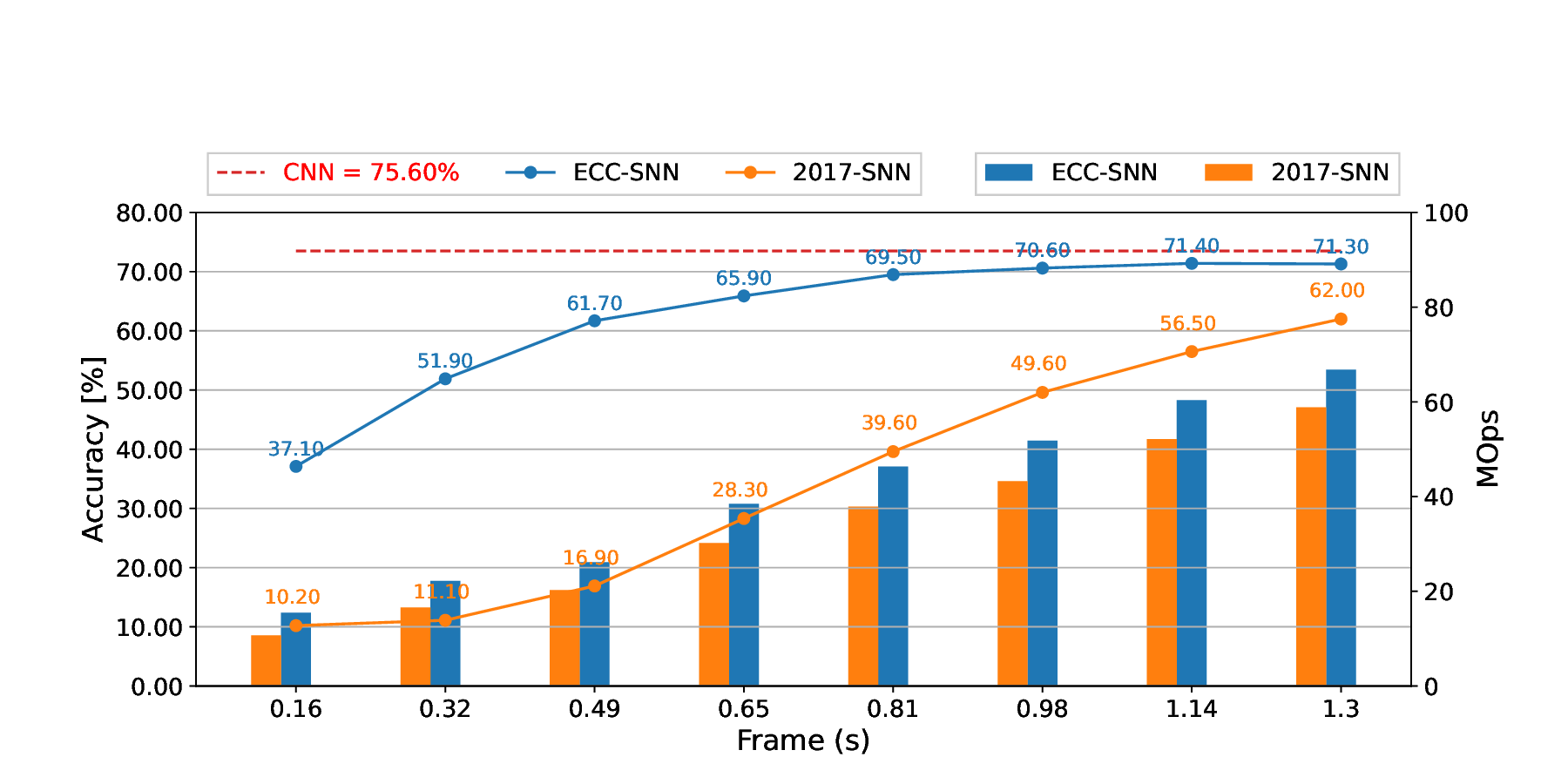}
    \caption{Accuracy and energy consumption (MOps) w.r.t. frames, between 2017-SNN and ECC-SNN, for CIFAR-10-DVS and VGG-16}
    \label{fig:dvsvgg16}
    \end{subfigure}
    \caption{Results for DVS datasets}
    \label{fig:dsvresults}
\end{figure*}
\section{Experiment on VGG-19 Architecture}

The following experiment on CIFAR-100 and VGG-19 shows that 80 timesteps has been sufficient for ECC-SNN  to achieve near-zero accuracy loss. On the other hand, since VGG-19 is 
deeper than VGG-16, the  accuracy loss of 2017-SNN \cite{Rueckauer:2017} becomes more  significant (See Figure \ref{fig:vgg19a}).

Moreover, we note that, comparing to 2017-SNN, ECC-SNN only has a very minor energy consumption increase, but ECC-SNN can achieve near-zero accuracy loss with much smaller timesteps (See Figure \ref{fig:vgg19b}). That is, ECC-SNN can achieve similar accuracy performance by using much less spike operations and shorter latency. \emph{These observations confirm the results we have for VGG-16 architecture}.

We remark that, we do not compare with \cite{Han:2020} (i.e., RMP-SNN) and \cite{Sengupta:2019} because both of them do not work with bath-normalisation layers, and without batch-normalisation layers, it is hard to train a high performance network.


\section{How $\eta$ affects the Number of Spikes}

Figure~\ref{fig:spikecount} presents the spike count of hidden layers, for 2017-SNN and ECC-SNN with two different $\eta$, for CIFAR-100. We note that, on the shadow layers, the spike count increase of ECC-SNN over 2017-SNN is minor. It becomes more significant for  deeper layers. The increased spikes help ECC-SNN to  follow the expecting distribution more closely. The expectation curve is extracted by encoding CNNs activation in each layer directly, which is without impact of $\Delta_i^n$ and represents the optimal case. When increasing $\eta$ from 1/2 to 2/3, the spike count distribution deviates further from its expectation.